%% file: egpaper_final.tex
\documentclass[10pt,twocolumn,letterpaper]{article}

\usepackage{iccv}
\usepackage{times}
\usepackage{epsfig}
\usepackage{graphicx}
\usepackage{amsmath}
\usepackage{amssymb}
\usepackage{multirow}
\usepackage{booktabs}

\usepackage[breaklinks=true,bookmarks=false]{hyperref}
\usepackage{url}

\iccvfinalcopy 

\def\iccvPaperID{****} 

\ificcvfinal\fi

\begin{document}

\title{4D Myocardium Reconstruction with Decoupled Motion and Shape Model}

\author{Xiaohan Yuan \hspace{2mm}\hspace{5mm} 
Cong Liu \hspace{2mm}\hspace{5mm} 
Yangang Wang\thanks{Corresponding author. E-mail: yangangwang@seu.edu.cn. All the authors from Southeast University are affiliated with the Key Laboratory of Measurement and Control of Complex Systems of Engineering, Ministry of Education, Nanjing, China. This work was supported in part by the National Natural Science Foundation of China (No. 62076061), the Natural Science Foundation of Jiangsu Province (No. BK20220127).}\hspace{2mm}\hspace{5mm} \\%
\\
Southeast University, China\\
}


\affiliation{Southeast University, China}

\maketitle

\ificcvfinal\thispagestyle{empty}\fi

\def\supmat{{\color{red}{Sup.~Mat~}}}

\begin{abstract}
   Estimating the shape and motion state of the myocardium is essential in diagnosing cardiovascular diseases.
   However, cine magnetic resonance (CMR) imaging is dominated by 2D slices, whose large slice spacing challenges inter-slice shape reconstruction and motion acquisition.
   To address this problem, we propose a 4D reconstruction method that decouples motion and shape, which can predict the inter-/intra- shape and motion estimation from a given sparse point cloud sequence obtained from limited slices. Our framework comprises a neural motion model and an end-diastolic (ED) shape model. The implicit ED shape model can learn a continuous boundary and encourage the motion model to predict without the supervision of ground truth deformation, and the motion model enables canonical input of the shape model by deforming any point from any phase to the ED phase. Additionally, the constructed ED-space enables pre-training of the shape model, thereby guiding the motion model and addressing the issue of data scarcity.
   We propose the first 4D myocardial dataset as we know and verify our method on the proposed, public, and cross-modal datasets, showing superior reconstruction performance and enabling various clinical applications.
\end{abstract}

\input{Introduction.tex}

\input{Relatedwork.tex}
\input{Method.tex}
\input{Experiments.tex}


\section{Conclusion}
The diagnosis of cardiovascular diseases depends on the accurate estimation of the shape and motion state of the myocardium, while the sparse information of the CMR slices brings challenges to the task of inter-slice reconstruction.
To address this problem, we propose a 4D myocardium reconstruction method, which can predict the complete shape and realize dense motion estimation from a given sparse point cloud sequence. 
We introduce the first 4D myocardial dataset, and evaluate our method on different datasets. We hope that our dataset and method can bring help and new insights to the field of 4D myocardium reconstruction.

\noindent
\textbf{Limitations and future work.}  
Our experiments show that our motion network is capable of learning the motion patterns that correspond to the heart's movement. However, we have not yet imposed explicit restrictions on the distribution of these motion patterns.
While the primary focus of this work is on the 4D reconstruction of healthy hearts, we also aim to validate our method using a dataset containing cardiac diseases (ACDC 2017 dataset). The experimental findings reveal the potential for improvement, particularly in cases with significant morphological disparities, such as dilated cardiomyopathy. Nevertheless, we maintain that our framework, as a general method, can be extended to encompass abnormal hearts given a broader distribution of data. The current progress substantiates the value of applying implicit decoupled representations in the cardiovascular domain, showcasing promising results in myocardial reconstruction. In the future, we plan to explore use of dense inter-slice motion representations in various applications, including the analysis of their correlation with disease incidence.


{\small
\bibliographystyle{ieee_fullname}
\bibliography{mybibliography}
}

\end{document}

%% file: Introduction.tex
\section{Introduction}
\begin{figure}[t]
    \centering
    \includegraphics[width=0.9\linewidth]{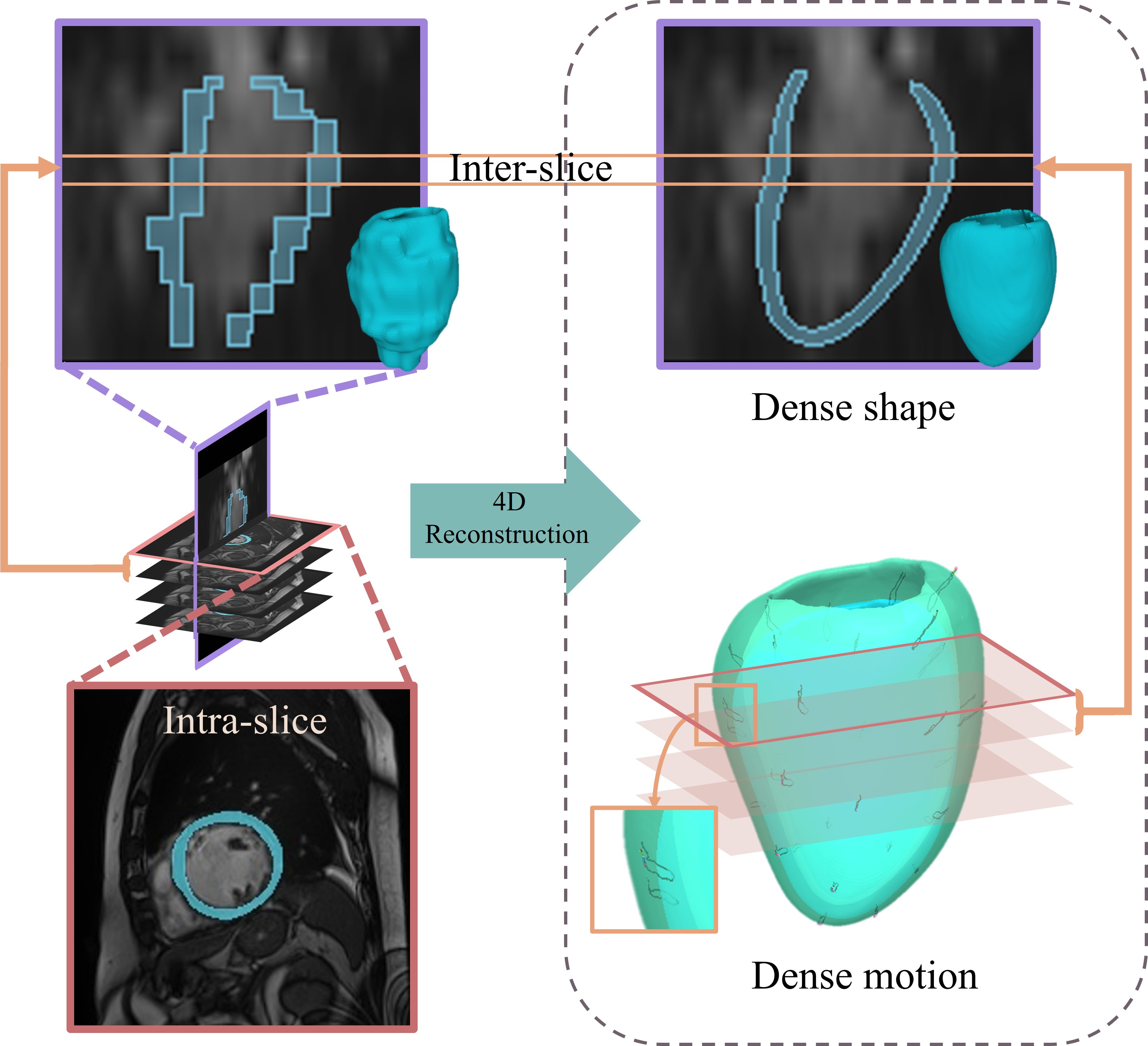}
    \caption{We propose a 4D cardiac reconstruction method predicting the inter-/intra- shape and motion of a cardiac cycle.}
    \label{teaser}
    \vspace{-5mm}
\end{figure}

Heart attack, also known as myocardial infarction, is the leading cause of death worldwide~\cite{who.int}. Estimating the shape and motion state of the myocardium is an essential step in diagnosing cardiovascular diseases. 
Cardiac magnetic resonance (CMR) imaging, which can provide successive sequence images including end-diastolic (ED) and end-systolic (ES) phases, is considered the gold standard for cardiac morphological assessment. 

However, the standard CMR has a large slice spacing, which leads to the lack of \textbf{inter-slice} information as shown in Fig.~\ref{teaser}. 
Meanwhile, the shape and motion observation of the ventricle for clinicians are almost guided by high-resolution two-dimensional slices based on \textbf{intra-slice} information, such as short-axis (SAX) or long-axis (LAX) views. The dense and faithful 4D (inter-/intra-) reconstruction of the myocardium would assist medical professionals in intuitively observing and accurately obtaining diverse clinical parameters~\cite{amzulescu2019myocardial}.

The road to inter-slice reconstruction is particularly challenging, as it involves estimating both inter-slice motion and shape. 
The lack of inter-slice motion data limits the applicability of supervised methods, thus complicating the problem. 
The large slice thickness of CMR leads to the inaccuracy of direct linear interpolation of pixels between slices, hindering the accurate reconstruction of the inter-slice shape. 
Few works have considered or fully addressed this problem.
Most recent works \cite{Yu2020MotionPN, Qin2020BiomechanicsinformedNN, Ye2021DeepTagAU, Zhang2022LearningCO} focus only on motion estimation based on 2D plane registration, ignoring the motion between slices and failing to reflect the myocardium movement truly. 
Others focus solely on 3D shape reconstruction of a specific phase \cite{attar20193d, Kong2021ADA, Joyce2022RapidIO}, without considering motion information. A few methods that attempt simultaneous shape and motion reconstruction \cite{Guo2021UnsupervisedLD, Meng2022MulViMotionS3} have limited resolution and cannot obtain dense point correspondence due to the complexity of cardiac motion.


In this work, we advocate to solve the inter-slice and intra-slice reconstruction as 4D myocardium reconstruction. Our key idea is to \textbf{decouple the problem as two steps including motion and shape reconstruction with sparse point clouds as inputs obtained from sequential CMR slices}. 
Inspired by the implicit neural representation~\cite{Park2019DeepSDFLC, Mescheder2019OccupancyNL, Deng2021DeformedIF, Zheng2021DeepIT},
we follow the point sampling strategy for both the motion model and shape model.
According to the characteristics of the heart sequence, we build the meaningful \textbf{E}nd-\textbf{D}iastolic\textbf{-space} (ED-space) that consists of a statistical parameter space representing the shape prior and a canonical spatial space suitable for model input. This space enables us to pre-train a reliable implicit shape model, which can learn a continuous boundary and guide the implicit motion model to obtain a dense motion field without the supervision of ground truth deformation.


Specifically, for any query point under a certain phase, the motion model takes its coordinate and phase indicator as input, estimating the deformation of the point to the ED spatial space under the condition of the motion code. 
The pre-trained shape model estimates the signed distance function (SDF) value of the deformed point under the condition of shape code in the ED spatial space. Finally, the shape reconstruction is completed by extracting the boundary (where SDF value = 0). 
Our method can be applied to multiple medical image analysis tasks such as point cloud completion, dense motion estimation, and motion interpolation. Furthermore, it is possible to extend our method to other modes of medical imaging, such as CT.

The main contributions of this paper are summarized as follows:

\noindent$\bullet$ We propose a new 4D myocardium reconstruction framework via the decoupled motion and shape features. 

\noindent$\bullet$ We build a ED shape model based on ED-space to alleviate the problem of medical data scarcity.

\noindent$\bullet$ We present the first, to our knowledge, 4D myocardial dataset composed of 3D shape sequence and the dataset can be downloaded from \url{https://github.com/yuan-xiaohan/4D-Myocardium-Reconstruction-with-Decoupled-Motion-and-Shape-Model}.

%% file: Relatedwork.tex
\section{Related work}

\begin{figure*}[ht]
    \centering
    \includegraphics[width=1.0\linewidth]{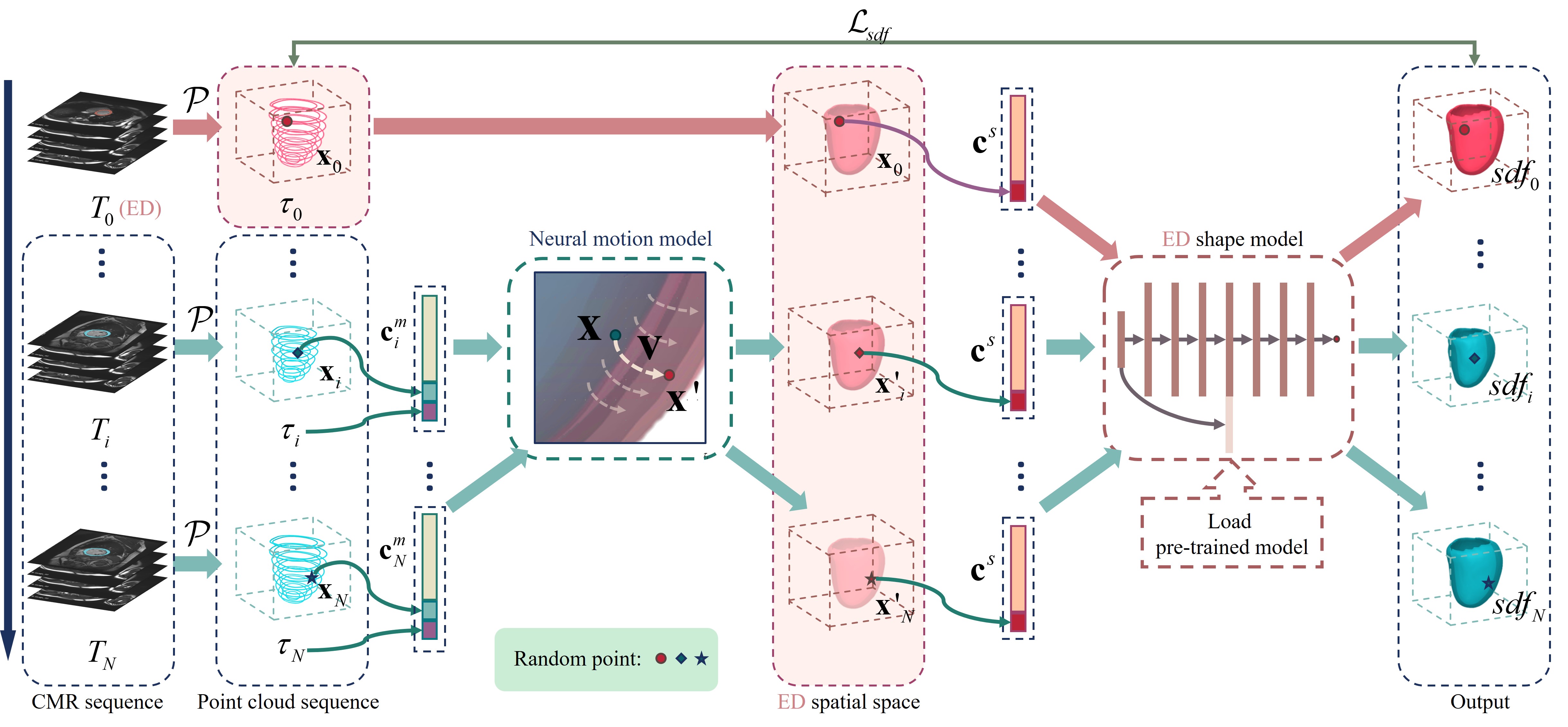}
    \caption{\textbf{Method Overview.} The point cloud sequence is first segmented from the input CMR sequence. We register the point cloud at the ED phase to the pre-defined statistical mean shape $\mathbf{\bar{s}}$ in the ED spatial space and broadcast the obtained transformation to all phases. For any query point at $T_{i}$ phase, the neural motion model takes its coordinate and phase indicator as input and predicts the point's deformation to the ED spatial space under the condition of motion code. The pre-trained implicit ED shape model estimates the SDF value of the deformed point under the condition of shape code. Finally, the shape reconstruction is completed by extracting the boundary.}
    \label{pipeline}
\end{figure*}
\subsection{Implicit representation}
Since the signed distance function (SDF) is a continuous function, implicit functions can represent more detailed and complicated information than mesh-based shape representation and are suitable for generating models \cite{Park2019DeepSDFLC}. In recent years, combining SDF with deep learning to represent shape has attracted broad interest \cite{Hao2020DualSDFSS, Deng2021DeformedIF, Zheng2021DeepIT}. Occupancy Flow extends 3D occupancy functions into the 4D domain by assigning a motion vector to every point in space and time \cite{Niemeyer2019OccupancyF4}.
By taking advantage of continuity of representation and delicate reconstruction, spatio-temporal implicit functions have been widely used in dynamic humans and clothing modeling \cite{Corona2021SMPLicitTG, Mihajlovi2021LEAPLA, Tiwari2021NeuralGIFNG, Palafox2021NPMsNP, Jiang2022LoRDL4}. However, there is few attempt to work on medical imaging at present. 
The application of implicit function in the medical field still has great potential to be tapped. DISSMs \cite{Raju2022DeepIS} present a new approach that marries the representation power of deep networks with the benefits of SSMs by using an implicit representation. ImplicitAtlas \cite{Yang2022ImplicitAtlasLD} and NDF \cite{Sun2022TopologyPreservingSR} both learn deep implicit shape templates. The former integrates multiple templates and represents shapes by deforming one of several learned templates. The latter represents the shape as the conditional differential deformation of the template to maintain the shape topology. Only static organs are considered in these works, while cardiac movement needs to consider time-related deformability.

The decoupling of motion and shape codes sets our method apart from other implicit methods and the key benefits include 
(1) Easier learning and understanding of features related to motion and shape; 
(2) Better utilization of diverse modal data (CMR and CT share a consistent shape code space; motion knowledge learned from CMR can be interpolated for CT);
(3) The decoupled shape codes offer advantages in pre-training the shape model. We define an interpretable canonical space, representing the spatial space at the ED phase, which is distinct from the canonical space used in template-based methods like DIT \cite{Zheng2021DeepIT}. This allows us to employ a data augmentation strategy to pre-train the ED shape model, thereby compensating for the data scarcity, as the training data is derived from the existing readily sampled statistical parameter space.

\subsection{Medical shape reconstruction}
Shape reconstruction methods for the medical image can be divided into parametric-based and deformation-based methods. The former needs to use a large amount of data to build the parameter space. It relies on accurate registration, mainly focuses on iterative optimization, and requires a long reference time \cite{frangi2002automatic, hoogendoorn2012high, bai2015bi}. With the development of deep learning, the combination of SSM and convolutional neural networks (CNNs) can achieve better results \cite{attar20193d, zhou2019one, adams2020uncertain}. However, the lack of medical image data is still the main reason for limiting such methods.
The deformation-based method can directly predict the surface mesh of organs from image or volume data \cite{mandikal20183d, ye2020pc, 8809843}. In recent years, the graph convolution neural network also shows the prospect of surface reconstruction \cite{wickramasinghe2020voxel2mesh, kong2021whole}. This kind of structured representation needs to build a fixed-topology template, which is difficult to obtain dense correspondences and is easily limited by the complexity of the shape.
In addition, the reconstruction of complete shapes from sparse point clouds from MR, namely point cloud completion, is also a problem worth studying \cite{Beetz2021BiventricularSR, Chen2021ShapeRW}.

\subsection{Cardiac motion estimation}
Motion estimation is to establish a corresponding deformation field between two phases, and image-based registration combined with deep learning is a more effective method at present. VoxelMorph \cite{Balakrishnan2019VoxelMorphAL} is a classic unsupervised registration network in medical images, which has served as a foundation for many follow-up studies aimed at its enhancement.
Most of the current work is about 2D motion reconstruction \cite{qin2018joint, krebs2019probabilistic, Qin2020BiomechanicsinformedNN, yu2020foal, chen2020anatomy, yu2020motion, Zhang2022LearningCO}.  
Other methods \cite{Guo2021UnsupervisedLD, meng2022mesh, Meng2022MulViMotionS3} are carried out on a 3D motion reconstruction. Based on the idea of dense-sparse-dense (DSD), Guo \textit{et al.} \cite{Guo2021UnsupervisedLD} extract sparse landmarks from the original image, then calculate the sparse displacement according to the detected sparse landmarks. Finally, they project the sparse motion displacement back to the dense image domain through the motion reconstruction network. However, this motion estimation accuracy will increase with the number of landmarks, and the detailed structural changes cannot be characterized when the number of landmarks is small. In order to solve the problem of cross-plane motion, a 3D motion field is estimated in \cite{Meng2022MulViMotionS3} by fusing the information of LAX and SAX views in latent space, and a shape regularization module is introduced to constrain the consistency of the 3D motion field. Meng \textit{et al.} \cite{meng2022mesh} also model the heart as a 3D geometric mesh and estimate the 3D motion of the cardiac mesh from the multi-view images. However, mesh-based representation requires dense vertices to represent the complex surface details of the heart.

%% file: Method.tex
\begin{figure*}[ht]
    \centering
    \includegraphics[width=0.9\linewidth]{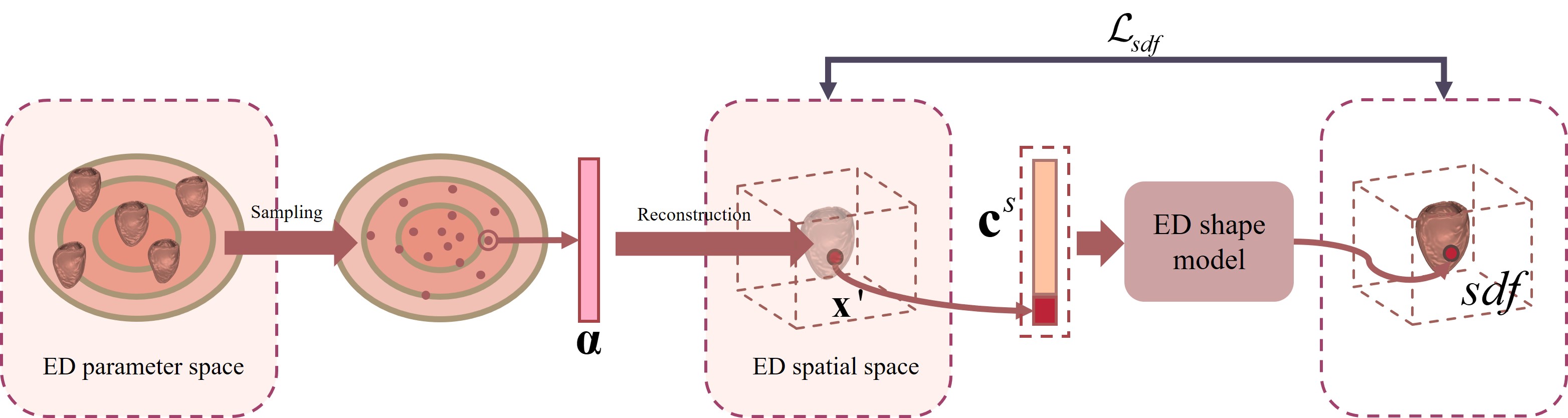}
    \caption{\textbf{Pipeline for ED shape model pre-training.} A shape parameter vector is sampled from the ED parameter space to reconstruct a new shape. This reconstructed shape is then normalized into the ED spatial space, and the pre-trained ED shape model is obtained by sampling points within the ED spatial space.}
    \label{EDspace}
\end{figure*}

\section{Method}
The overview of our proposed method is shown in Fig.~\ref{pipeline}. 
We decouple the task into a two-step subproblem, including motion reconstruction and ED shape reconstruction, where motion reconstruction focuses on deforming the point at any phase into the ED phase, and ED shape reconstruction can facilitate the reconstruction loss computation.
It is noted that the raw input CMR slices might have different directions and positions due to the diversity of data acquisition. Point clouds should be pre-processed and registered. 
At the beginning, we register the point cloud at the ED phase with the pre-defined statistical mean shape $\mathbf{\bar{s}}$ to compute a $4\times4$ matrix $\mathcal{P}$ (details are shown in \supmat).
After that, $\mathcal{P}$ is applied to the whole sequence to transform all point clouds (shown in Fig.~\ref{pipeline}) as the sampling input for the neural motion model.
For any query point at $T_{i}$ phase, the motion model (Sec.~\ref{Implicit motion model}) predicts its deformation relative to that point in ED-space (Sec.~\ref{ED-space building}) under the condition of the motion code. Then, the pre-trained ED shape model (Sec.~\ref{Implicit shape model}) estimates the SDF value $sdf$ under the condition of the shape 
code. Finally, we can easily obtain the myocardium reconstruction mesh via Marching Cubes~\cite{lorensen1987marching}. Before we train the framework, the constructed ED parameter space is sampled for data augmentation to improve the generalization performance of the shape model.

\subsection{Neural motion model}
\label{Implicit motion model}
A neural motion model is formulated to predict the deformation of a given point at arbitrary $T_{i}$ phase to the end-diastole (ED) phase (\ie, $T_{0}=0$). To construct the motion model, all points should be registered and sampled in a canonical space. We align the input ED phase point clouds with a statistical mean shape $\mathbf{\bar{s}}$ defined in Sec.~\ref{ED-space building} and apply the obtained transformation matrix $\mathcal{P}$ to all phases.
For a sampling point $\mathbf{x}\in \mathbb{R}^{3}$ in the canonical space, its corresponding deformation $\mathbf{v}$ is predicted under the condition of motion code $\mathbf{c}^{\textrm{m}}_{i} \in \mathbb{R}^{K^{\textrm{m}}}$ and phase indicator $\tau_i$:
\begin{equation}
    f^{\textrm{m}}:(\mathbf{x}, \mathbf{c}^{\textrm{m}}_{i}, \tau_i)\in \mathbb{R}^{1+3+K^{\textrm{m}}}\longrightarrow \mathbf{v}\in \mathbb{R}^{3},
\end{equation}
where $K^{\textrm{m}}$ is the dimension of the motion code. $\tau_i=T_i/T_{N} \in [0,1]$, $T_{i}$ is the $i$-th phase of CMR sequence, $T_{N}$ is the sequence length, and $T_{0} = 0$ is for ED phase. It is noted that the motion code $\mathbf{c}^{\textrm{m}}_{i}$ is variant for each $i$-th phase. With the proposed neural motion model, the corresponding point $\mathbf{x'}$ in ED spatial space can be obtained by $\mathbf{x'}=\mathbf{x}+\mathbf{v}$. Our proposed motion model can obtain the deformation from any point at the $T_{i}$ phase to the ED phase; thus, we can easily establish dense correspondences. To train the motion model in an unsupervised manner, we utilize the reconstruction loss to compare the differences between the input point cloud and the output of the shape model. 
In order to better guide the motion model, it is crucial to have accurate SDF value predictions. Next, we will incorporate a pre-trained shape model. 

\subsection{ED shape model}
\subsubsection{ED-space building}
\label{ED-space building}
As the first frame of the heart sequence typically represents the pivotal ED phase, we select this frame as the reference. Based on the unique characteristics of the heart sequence, we construct the ED-space that consists of a statistical parameter space representing the shape prior and a canonical spatial space dedicated to the heart as the input space of the shape model (as shown in Fig.~\ref{EDspace}).

To build the ED parameter space, we introduce the left ventricular ED statistical shape model (SSM) \cite{bai2015bi}, obtained from high-resolution MR images of more than 1,000 healthy subjects, describing individual shape change. For the given shape parameter vector $\mathbf{\alpha}\in \mathbb{R}^{K^{\alpha}}$, we can get the corresponding 3D shape through the following function:
\begin{equation}
    \mathbf{S}=\mathcal{M}(\mathbf{\bar{s}}+\mathbf{V^{\textrm{s}}}\mathbf{\alpha}),
    \label{pca}
\end{equation}
where $K^{\alpha}$ is the dimension of the parameter vector, $N$ is the number of vertices, $\mathbf{V^{\textrm{s}}} \in \mathbb{R}^{3N\times K^{\alpha}}$ is obtained through the singular value decomposition (SVD) of the atlas map, $\mathbf{\bar{s}}$ is the statistical mean shape, and the operator $\mathcal{M}(\cdot):\mathbb{R}^{3N}\mapsto \mathbb{R}^{3\times N}$ maps the parameter vector to the 3D shape.
To improve the training of the ED shape model, we generate a multivariate Gaussian distribution centered on real samples for sampling to augment. The newly obtained parameters differ from the original atlas while maintaining anatomical consistency and fidelity. 
All new shapes obtained from Eq.~\ref{pca} are naturally registered with the mean shape $\mathbf{\bar{s}}$, and they can be directly normalized into a canonical space called ED spatial space. At the beginning of the whole framework, all our point cloud inputs will be aligned to this space for sampling.

\subsubsection{Shape model pre-training}
\label{Implicit shape model}
To improve the accuracy of the ED shape model, we use a large number of shapes obtained through the above data augmentation strategy to pre-train an implicit shape function for the ED-space. As shown in Fig.~\ref{EDspace}, for a query point $\mathbf{x}\in \mathbb{R}^{3}$ in the ED spatial space, the corresponding SDF value $sdf$ can be predicted through an MLP network based on the shape code $\mathbf{c}^{\textrm{s}}\in \mathbb{R}^{K^{\textrm{s}}}$:
\begin{equation}
    f^{\textrm{s}}:(\mathbf{c}^{\textrm{s}}, \mathbf{x})\in \mathbb{R}^{K^{\textrm{s}}+3}\longrightarrow sdf\in \mathbb{R}.
\end{equation}
where $K^{\textrm{s}}$ is the dimension of the shape code. The implicit shape model learns the shape code $\mathbf{c}^{\textrm{s}}$ of a given object, and the weight of network $f^{\textrm{s}}$ in the fashion of automatic decoder \cite{park2019deepsdf} simultaneously. Note that $\mathbf{c}^{\textrm{s}}$ is shared in the same sequence for all the points. After querying the predicted $sdf$ of each vertex in the spatial grid, the decision boundary of zero isosurfaces is extracted. The final mesh is obtained using Marching Cubes \cite{lorensen1987marching}. The obtained pre-trained model will be applied to the joint framework for fine-tuning.

\subsection{Loss functions}
Given the shapes of a motion sequence, the SDF regression loss is applied:
\begin{equation}
    \mathcal{L}_{sdf}= \sum_{i=1}^{S}\sum_{l=1}^{T_{N}} \sum_{j\in {\Omega} }\left | f(\mathbf{x}_{j})-sdf_{j} \right |,
\end{equation}
where $S$ and $T_{N}$ represent the number and the length of sequences respectively, ${\Omega}$ denotes points in 3D space, $f=f^{\textrm{s}}(f^{\textrm{m}}(\cdot))$ describes the process of point $\mathbf{x}_{j}$ passing through the motion model and the ED shape model in turn, and $sdf_{j}$ is the ground-truth SDF value. This loss also provides vital guidance for the neural motion model.

In addition, We use the regularization terms \cite{Zheng2021DeepIT} $\mathcal{L}_{pw}$ and $\mathcal{L}_{pp}$ to prevent unrealistic large distortion:
\begin{equation}
    \mathcal{L}_{pw}=\sum_{i=1}^{S} \sum_{l=1}^{T_{N}} \sum_{j\in \Omega } h(\left \|f^{\textrm{m}}(\mathbf{c}^{\textrm{m}}_{l}, \mathbf{x}_{j}, \tau_{j})-\mathbf{x}_{j}  \right \|_{2}),
\end{equation}
where $h(\cdot )$ is the Huber kernel, and
\begin{equation}
    \mathcal{L}_{pp}=\sum_{i=1}^{S} \sum_{l=1}^{T_{N}} \sum_{j, k\in \Omega,j\ne k} max(\frac{\left \| \hat{\mathbf{v}}_{j}-\hat{\mathbf{v}}_{k} \right \|_{2} }{\left \| \mathbf{x}_{j}-\mathbf{x}_{k} \right \|_{2} }-\epsilon ,0 ),
\end{equation}
where $\epsilon$ is a parameter controlling the distortion tolerance.

Finally, the total training loss is expressed as:
\begin{equation}
    \mathcal{L} = \omega_{1}\mathcal{L}_{sdf} + \omega_{2}\mathcal{L}_{pw} + \omega_{3}\mathcal{L}_{pp} + \omega_{4}\mathcal{L}_{reg},
\end{equation}
where $\mathcal{L}_{reg}=\left \| \mathbf{c}^{\textrm{m}} \right \|_{2} + \left \| \mathbf{c}^{\textrm{s}} \right \|_{2}  $ is the regularization term of latent codes, and $\omega_{\cdot}$ are the weights for different loss terms.

\subsection{Optimization as inference}
After training, we can reconstruct the 4D shape based on the given sparse CMR slice sequence. We manually delineate all the slices of each sequence and apply the segmentation network nn-Unet \cite{Isensee2019nnUNetBT} for training to obtain the contour point clouds of the left myocardium. Here we assume that the slice-level segmentations provided represent the surface of the myocardium, with reference SDF values of 0, and use their supervision to optimize the latent codes. We can use either the short-axis (SAX) or long-axis (LAX) views of CMR for segmentation, but due to the large slice spacing, the resulting point clouds are still sparse at the inter-slice level. Before entering the motion model, we obtain $\mathcal{P}$ by aligning the point cloud at the ED phase to the ED spatial space and applying it to the entire sequence. Thanks to the automatic decoder structure, we fix the network weight here and optimize the latent codes $\mathbf{c}^{\textrm{m}}$ and $\mathbf{c}^{\textrm{s}}$. Note that the same sequence shares the $\mathbf{c}^{\textrm{s}}$, and each phase has its own $\mathbf{c}^{\textrm{m}}$.


\noindent\textbf{Implementation details.}
Our neural motion model and ED shape model use a multi-slice perceptron structure. We choose $K^{\textrm{m}}=128$ and $K^{\textrm{s}}=256$ for the dimensions of codes. In the training set, $T_{N}=25$. The learning rate ($1e-3$) of the motion model is set to be greater than the learning rate ($1e-4$) of the shape model so that a good shape model can make the motion model learn and deform better. For each loss weight, we set $\omega_{1}=1$, $\omega_{2}=5e-3$, $\omega_{3}=1e-4$, $\omega_{4}=1e-4$. 

%% file: Experiments.tex
\section{Experiments}


\begin{figure}[!t]
    \centering
    \includegraphics[width=1.0\linewidth]{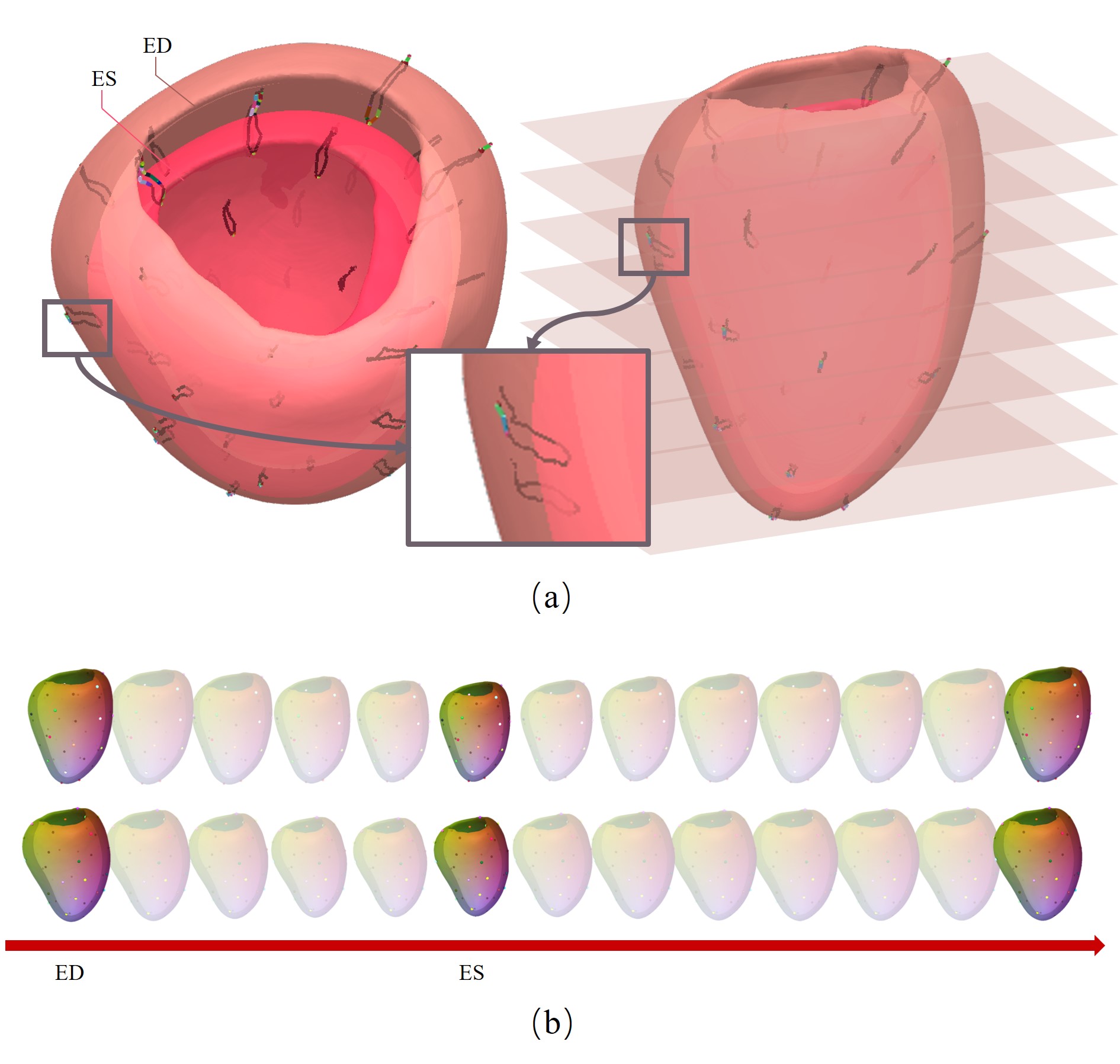}
    \caption{Dense motion estimation. (a) Presentation of dense motion estimation. Some sample points are highlighted. (b) The motion track of some surface points (black curve), one of which is zoomed in. The shapes of the ED phase and ES phase are visualized.}
    \label{motion}
\end{figure}

\noindent \textbf{Data.}
We train the model on our proposed CMR dataset, which was acquired at Jiangsu Province Hospital, and composed of 55 healthy subjects. Each subject includes multiple slices (8-10 slices) with a resolution of $1.25\times1.25\times10 mm$. Each slice covers the video sequence of the cardiac cycle (25 phases). The clinical experts manually delineated the left myocardium of all the phases and slices. More information are presented in \supmat.

All the models are trained only on our proposed dataset and evaluate on the proposed dataset, the publicly available ACDC 2017 dataset~\cite{bernard2018deep}, and the CT dataset. For the proposed internal dataset, we randomly split it into training, testing, and validation sets in a ratio of $60\%$, $20\%$, and $20\%$. The internal validation set is used for optimizing hyperparameters and held out until final evaluation. The ACDC 2017 dataset contains CMRs of 100 patients in five categories, each with a slice thickness of 5-10 mm and each sequence covering at least one cardiac cycle, with varying lengths. The CT dataset was collected in Jiangsu Province Hospital, and each only covers the two critical phases of ED and ES, which will verify the ability of our method to perform the cross-modal interpolation.

\noindent \textbf{Evaluation metrics.}
Our proposed dataset uses Chamfer Distance (CD) and Earth Mover Distance (EMD) to evaluate the shape similarity between the ground truth and predicted reconstruction. Since the ACDC 2017 and CT datasets have no ground truth shape, we use the Dice score (Dice) and Hausdorff distance (HD) to measure the segmentation results on the intra-slice instead.


\subsection{Applications}
\subsubsection{Dense motion estimation}
We select some surface points to visualize the motion track of the cardiac cycle, as shown in Fig.~\ref{motion} (a). It can be seen that the motion obtained by our method will not be confined to the slice. It will serve the task of 3D motion tracking well and analyze the motion state of key points. The presentation of dense motion estimation is shown in Fig.~\ref{motion} (b), where the consistent colors across different phases correspond to the same points. We can trace the corresponding point for any given point at any phase. The acquisition of this dense field holds immense potential for analyzing more statistical and clinical indicators.

\subsubsection{Motion interpolation}
\begin{figure}[!t]
    \centering
    \includegraphics[width=1.0\linewidth]{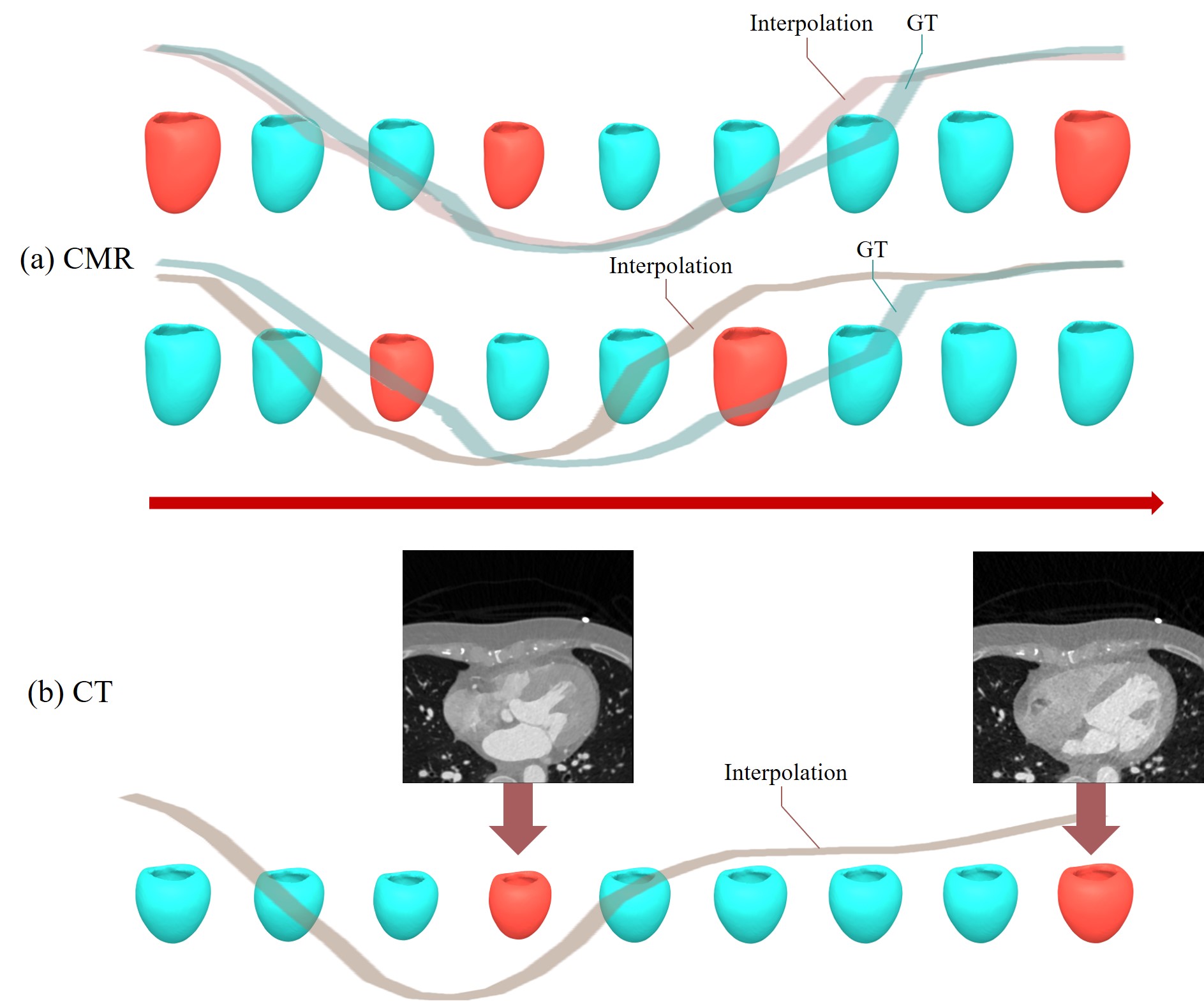}
    \caption{Visualization of motion code interpolation for CMR and CT. Cyan meshes are interpolated phases. The curves represent the corresponding volume change.}
    \label{interpolation}
\end{figure}

Because our method can decouple the motion and shape, we can estimate the missing shapes by interpolating the motion code from the incomplete sequence. We select $L$ ($L<T_{N}, T_{N}$ is the length of the complete sequence) phases as observations. 
We use PCA to extract features from the motion code map of the whole sequence from the training set $\mathbf{Seq}\in \mathbb{R}^{S\times T_{N}\cdot K^{\textrm{m}}}$ ($S$ is the number of the sequence), and get the motion parameter vector $\mathbf{\beta}\in \mathbb{R}^{K^{\mathbf{\beta}}}$ characterizing the sequence by regressing the observation:
\begin{equation}
    \underset{\mathbf{\beta} }{arg \mathrm{min}} \sum_{i=1}^{L} \left \| \mathbf{V}^{\textrm{m}}_{i}\mathbf{\beta}-\mathbf{c}^{\textrm{m}}_{i}\right \|_{2}^{2},
\end{equation}
where $\mathbf{V^{\textrm{m}}} \in \mathbb{R}^{T_{N}\cdot K^{\textrm{m}}\times K^{\beta}}$ is obtained through the SVD of $\mathbf{Seq}$, $\mathbf{V}^{\textrm{m}}_{i} \in \mathbb{R}^{K^{\textrm{m}}\times K^{\mathbf{\beta}}}$and $\mathbf{c}^{\textrm{m}}_{i}$ represent the $i$th observation of $\mathbf{V^{\textrm{m}}}$ and the motion codes respectively.

\begin{equation}
    \mathbf{M}=\mathcal{F}(\mathbf{\bar{m}}+\mathbf{V^{\textrm{m}}}\mathbf{\beta}),
    \label{pca_m}
\end{equation}
where $\mathbf{\bar{m}}$ is the mean motion codes of the sequence, and the operator $\mathcal{F}(\cdot):\mathbb{R}^{T_{N}\cdot K^{\textrm{m}}}\mapsto \mathbb{R}^{T_{N}\times K^{\textrm{m}}}$. $\mathbf{M}$ contains all motion codes of the entire sequence after completion.

We produce some reasonable results, as shown in Fig.~\ref{interpolation} (a). The first column gives keyframes (ED, ES, and the last frame) and interpolates the remaining phases. The second column randomly gives only two frames. The volume curve of the whole sequence is also displayed. It can be seen that the results obtained by our interpolation method can ensure the diastolic-systolic-diastolic of the heart cycle, reflecting the fidelity of motion code learning. 

Our application can also be adapted to work with CT images. We aim to convert CT data into a format that our model can handle, and leverage the prior knowledge acquired from analyzing CMR sequences to perform cross-modal motion interpolation. Typically, CT scans have a low temporal resolution and consist of only two keyframes, which is consistent with our assumption. By applying our method, we can reconstruct sparse CT data and generate a sequence with a similar temporal resolution to CMR scans, as shown in Fig.~\ref{interpolation} (b). This could be particularly useful in clinical settings where doctors need a reliable reference for monitoring motion.



\subsection{Ablation study on ED-space building}
\begin{figure}[!t]
    \centering
    \includegraphics[width=0.9\linewidth]{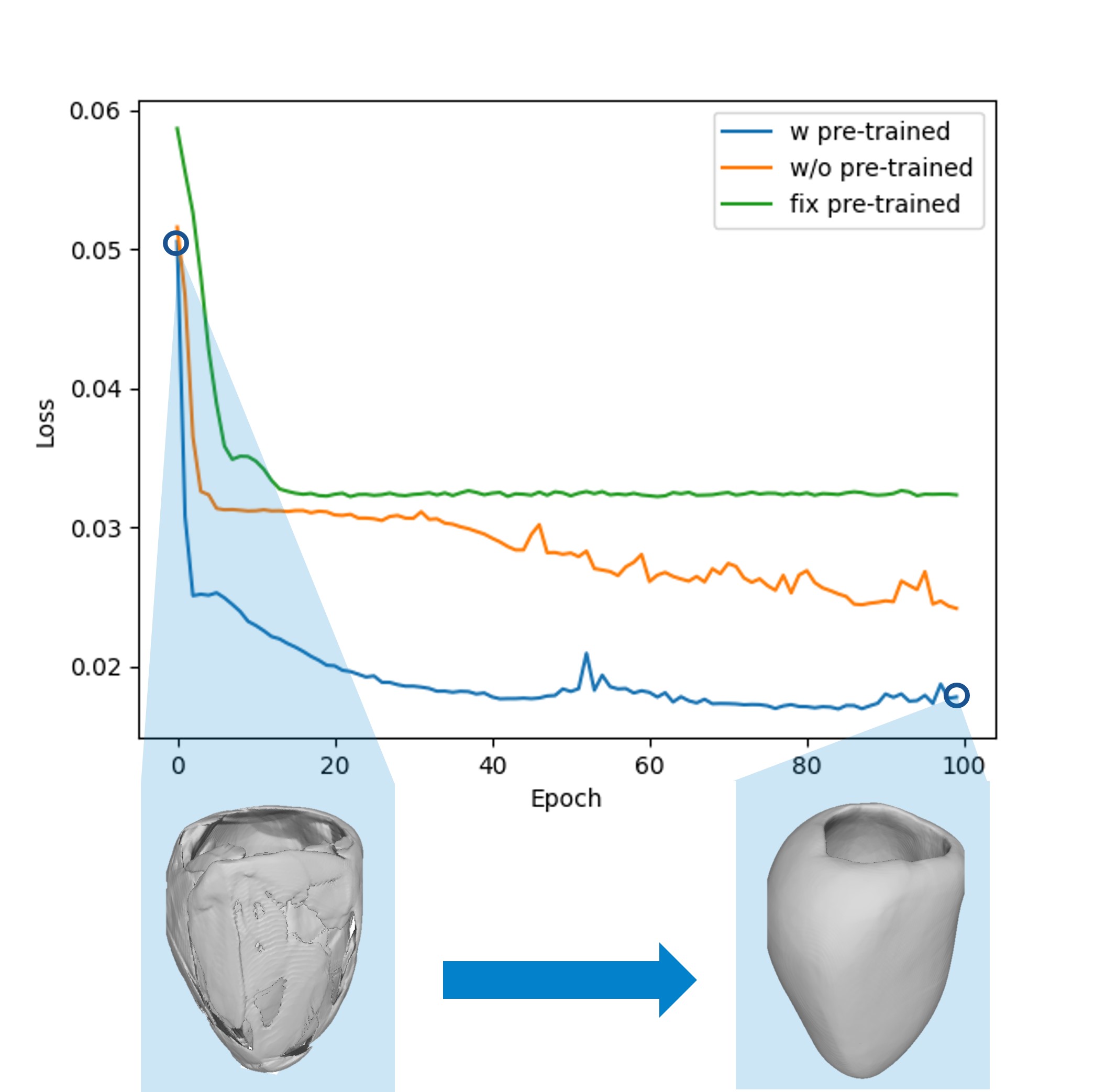}
    \caption{The influence of whether to establish ED-space in advance to pre-train the shape model.}
    \label{abla_losses}
\end{figure}
To demonstrate the effectiveness of pre-training the shape model by building the ED-space in advance, we draw the loss curve as shown in Fig.~\ref{abla_losses}. After constructing the statistical shape model, we sample in the parameter space to obtain many shapes to pre-train an implicit shape function. However, simply fixing the weight of the ED shape model and training only the motion model could result in the network becoming stuck in local optimization. Instead, if the trained shape model is put into the overall framework and trained together with the motion model, it can improve the accuracy of the ED shape model quickly. In our experiment, we set the learning rate of the motion model to be greater than that of the shape model so that a good shape model urges the motion model to learn and deform better. Fig.~\ref{abla_losses} shows an example where the pre-trained shape model cannot be well estimated initially, but the final inaccuracies will be corrected with the joint training.

\subsection{Comparisons}

\begin{figure}[!t]
    \centering
    \includegraphics[width=1.0\linewidth]{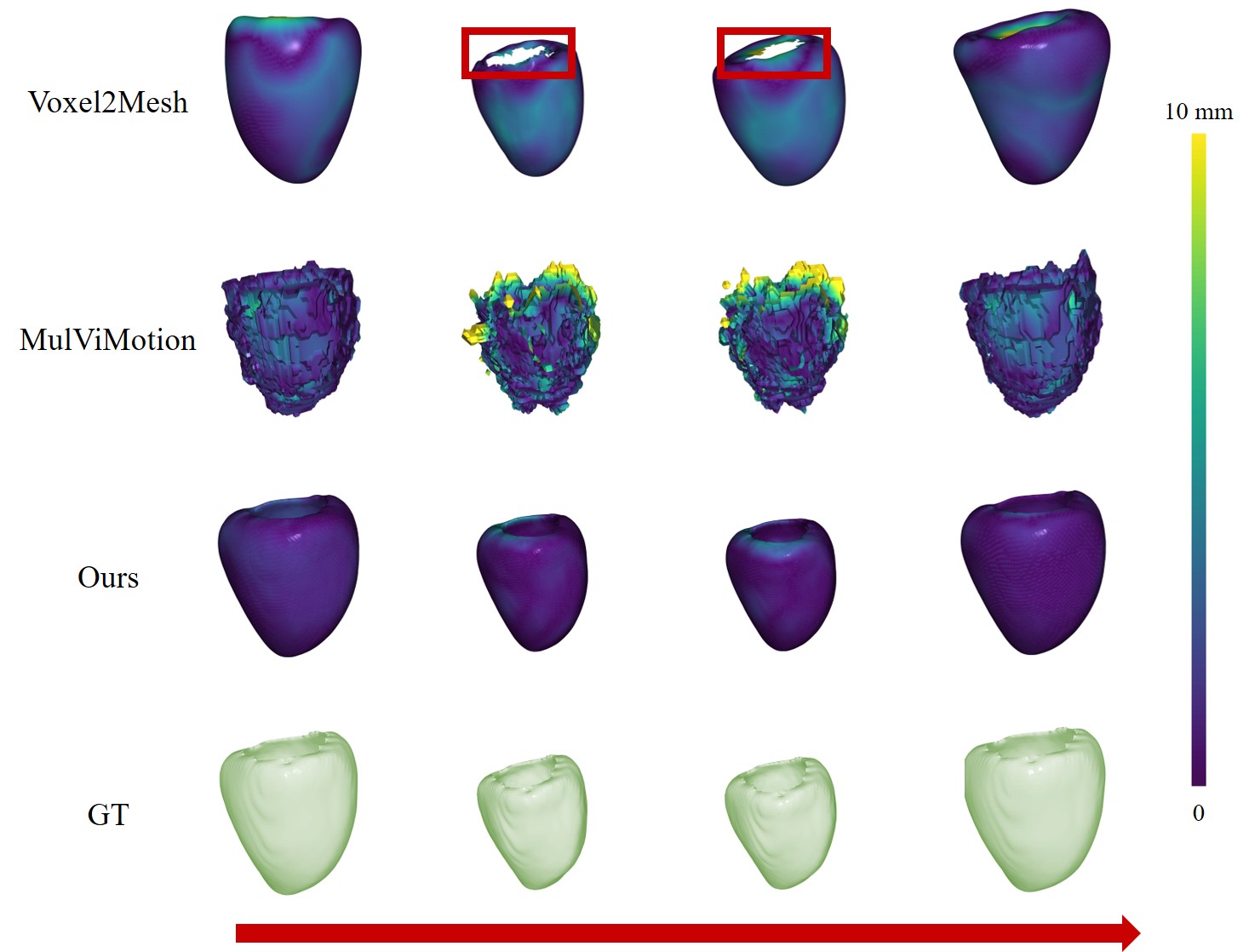}
    \caption{Error distribution of partial 4D reconstruction results of cardiac cycle based on explicit methods.}
    \label{comparison_explicit}
\end{figure}

\begin{figure*}[!t]
    \centering
    \includegraphics[width=1.0\linewidth]{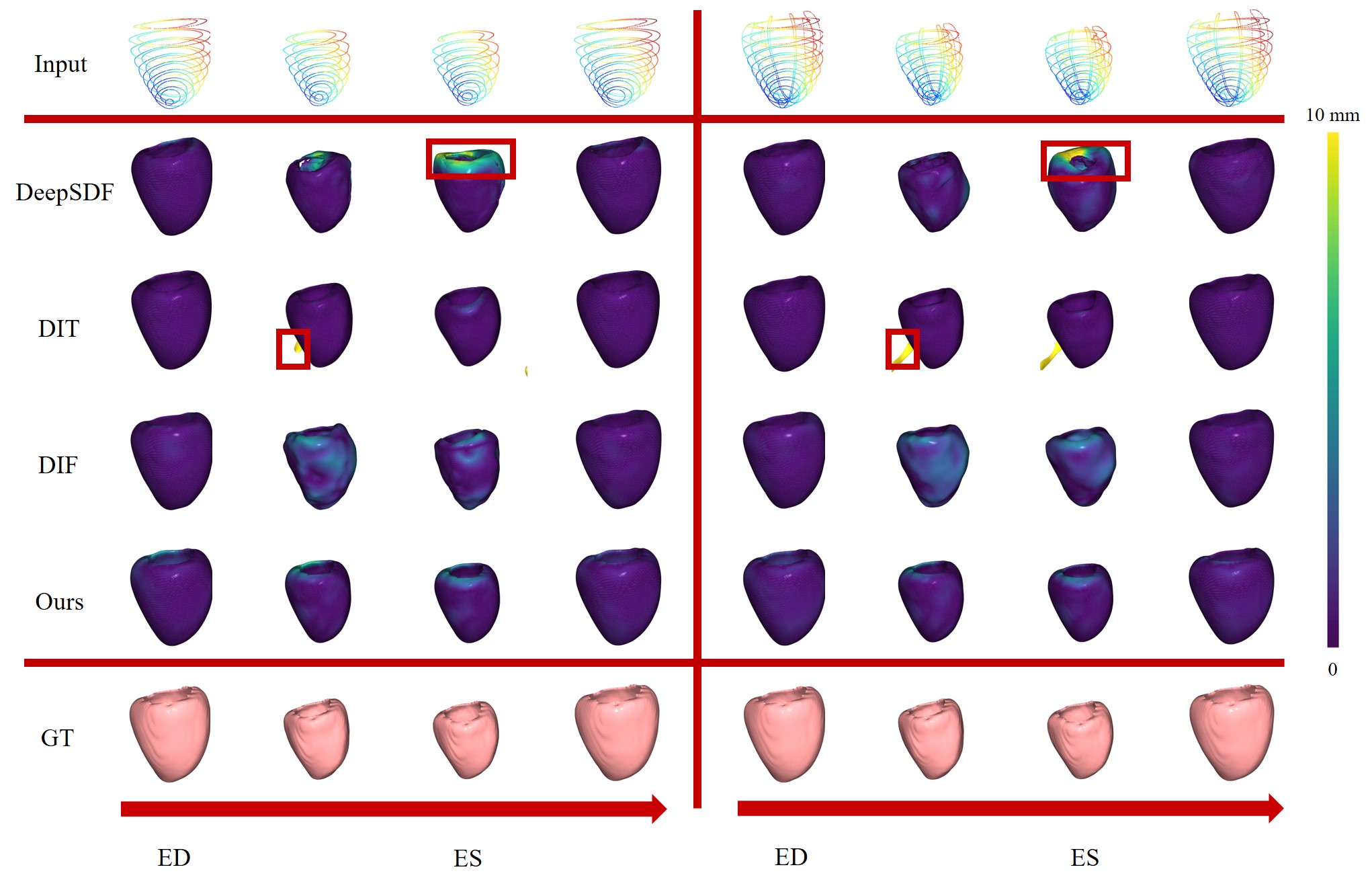}
    \caption{Error distribution of 4D reconstruction results of cardiac cycle based on implicit methods. The left uses the CMR SAX contours as input, and the right uses both the SAX and LAX contours as input. Our results are smoother and more fidelity in time sequence.}
    \label{comparison1}
\end{figure*}

\begin{table}[th]
    \caption{Quantitative results of 4D reconstruction on proposed dataset}
    \centering
    \begin{tabular}{cccc}
        \bottomrule
        Method      & Input views                    & CD$\downarrow$             & EMD$\downarrow$ \\ \hline
        Voxel2Mesh \cite{wickramasinghe2020voxel2mesh}  & Volume     & \multicolumn{1}{c}{22.842}                    & \multicolumn{1}{c}{7.796}     \\\hline
        DeepSDF \cite{Park2019DeepSDFLC}     &   \multirow{4}{*}{SAX}                     & \multicolumn{1}{c}{3.331} &      \multicolumn{1}{c}{6.447} \\
        DIT \cite{Zheng2021DeepIT}         &                          & \multicolumn{1}{c}{7.553} &    \multicolumn{1}{c}{7.645}                  \\
        DIF \cite{Deng2021DeformedIF}         &                          & \multicolumn{1}{c}{2.891} &     \multicolumn{1}{c}{7.010}\\
        Ours        &                          & \multicolumn{1}{c}{\textbf{2.634}} &    \multicolumn{1}{c}{\textbf{6.173}}  \\ \hline
        DeepSDF \cite{Park2019DeepSDFLC}    &   \multirow{5}{*}{SAX+LAX}                       &          \multicolumn{1}{c}{2.852}            &     \multicolumn{1}{c}{6.172}           \\
        DIT \cite{Zheng2021DeepIT}         &                          &          \multicolumn{1}{c}{8.156}            &      \multicolumn{1}{c}{6.878}               \\
        DIF \cite{Deng2021DeformedIF}         &                          &          \multicolumn{1}{c}{2.858}            &     \multicolumn{1}{c}{6.010}                   \\
        MulViMotion \cite{Meng2022MulViMotionS3} &                          &         \multicolumn{1}{c}{17.728}             &    \multicolumn{1}{c}{7.774}                 \\
        Ours        &                          &         \multicolumn{1}{c}{ \textbf{2.627}}             &   \multicolumn{1}{c}{ \textbf{3.603}}     \\ \bottomrule
    \end{tabular}
    \label{comparison_tab1}
\end{table}

\begin{figure*}[th]
    \centering
    \includegraphics[width=1.0\linewidth]{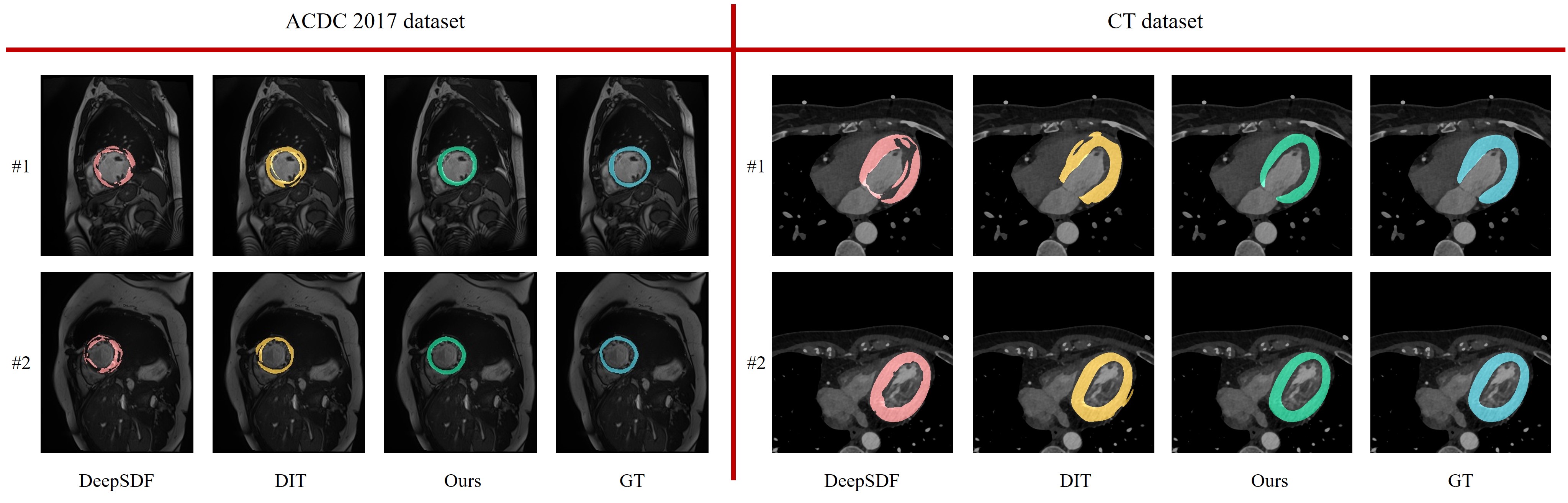}
    \caption{Segmentation results on the ACDC 2017 and CT datasets. We achieve smoother results that are more closely to the ground truth.}
    \label{comparison2}
\end{figure*}

\begin{table}[th]
    \caption{Quantitative segmentation results of 4D reconstruction on the ACDC 2017 dataset and CT dataset}
    \centering
    \begin{tabular}{cccc}
        \bottomrule
        Dataset & Method      & Dice$\uparrow$  & HD$\downarrow$ \\ \hline
        \multirow{4}{*}{ACDC 2017 dataset}
        & Voxel2Mesh  \cite{wickramasinghe2020voxel2mesh}    & \multicolumn{1}{c}{0.553} &      \multicolumn{1}{c}{3.534} \\
        & DeepSDF \cite{Park2019DeepSDFLC}    & \multicolumn{1}{c}{0.697} &      \multicolumn{1}{c}{3.257} \\
        & DIT \cite{Zheng2021DeepIT}         & \multicolumn{1}{c}{0.702} &      \multicolumn{1}{c}{3.011} \\
        & Ours        & \multicolumn{1}{c}{\textbf{0.765}} &      \multicolumn{1}{c}{\textbf{2.789}}  \\ \hline
        \multirow{3}{*}{CT dataset}
        & DeepSDF \cite{Park2019DeepSDFLC}    & \multicolumn{1}{c}{0.796} &      \multicolumn{1}{c}{3.048} \\
        & DIT \cite{Zheng2021DeepIT}         & \multicolumn{1}{c}{0.817} &      \multicolumn{1}{c}{2.646} \\
        & Ours        & \multicolumn{1}{c}{\textbf{0.842}} &      \multicolumn{1}{c}{\textbf{2.531}}  \\ \bottomrule
    \end{tabular}
    \label{comparison_tab2}
\end{table}

We investigate the representation power of different methods for 4D reconstruction. The quantitative results compared with the above methods are shown in Tab.~\ref{comparison_tab1} and Tab.~\ref{comparison_tab2}. 

\noindent \textbf{Comparison with explicit methods.}
For explicit methods, Voxel2Mesh \cite{wickramasinghe2020voxel2mesh} takes 3D medical volume as input and employs GCN. MulViMotion \cite{Meng2022MulViMotionS3} is designed for multi-view motion estimation of the heart and predicts the shape simultaneously. As shown in Fig.~\ref{comparison_explicit}, Voxel2Mesh loses many details and cannot reconstruct complex structures such as the opening and inner wall of the myocardium. MulViMotion has limited resolution and cannot reconstruct high-quality shapes. In contrast, our method utilizes implicit functions to obtain fine shapes and dense motion. 

\noindent \textbf{Comparison with implicit methods.} To ensure
fairness in the comparative experiments, the pre-processing steps of all implicit methods are consistent with our method, meaning that the input data is the same.
DeepSDF \cite{Park2019DeepSDFLC} is a classic structure without designing a deformation model. Both DIT \cite{Zheng2021DeepIT} and DIF \cite{Deng2021DeformedIF} are represented by an implicit template field, while their codes are not decoupled. By comparing them, we can prove the significance of decoupling motion and shape. We evaluate the results of taking only the CMR short-axis (SAX) segmentation contours as input and taking both the short-axis and the long-axis (SAX+LAX) segmentation contours as input on our dataset, and the error visualization is shown in Fig.~\ref{comparison1}. DeepSDF has good reconstruction results in the time phase close to ED. However, it can not make a good prediction in the systolic process with a large difference from ED, which shows that this method is not robust to reconstruct large deformation objects. DIF has the same problem, that is, it is difficult to accurately estimate systolic phases and cannot maintain a smooth topology. DIT can accurately reconstruct the shape of most phases, but there are many unexpected situations, as shown in the red box in Fig.~\ref{comparison1}, such as some additional parts. Unlike them, due to the separation of motion code, our method adds temporal information to further improve the continuity of motion, making the deformation smooth in time and the overall shape more reasonable. 

Fig.~\ref{comparison2} displays the segmentation outcomes for the ACDC 2017 dataset and CT dataset. Compared with other implicit methods, the segmentation we estimate on the intra-slice slice is closer to the ground truth. Our overall contour is smoother, and there are no discernible defects. According to Tab.~\ref{comparison_tab2}, due to limitations imposed by the types of training data, there is still considerable room for improvement in our method when applied to ACDC 2017 dataset with diseases.

%% file: egpaper_final.bbl
\begin{thebibliography}{10}\itemsep=-1pt

\bibitem{who.int}
Who cardiovascular diseases.
\newblock \url{https://www.who.int/cardiovascular_diseases/about_cvd/en/}.
  Accessed July 29, 2020.

\bibitem{adams2020uncertain}
Jadie Adams, Riddhish Bhalodia, and Shireen Elhabian.
\newblock Uncertain-deepssm: From images to probabilistic shape models.
\newblock In {\em International Workshop on Shape in Medical Imaging}, pages
  57--72. Springer, 2020.

\bibitem{amzulescu2019myocardial}
Mihaela~Silvia Amzulescu, Mathieu De~Craene, H{\'e}l{\`e}ne Langet, Agnes
  Pasquet, David Vancraeynest, Anne-Catherine Pouleur, Jean-Louis
  Vanoverschelde, and BL Gerber.
\newblock Myocardial strain imaging: review of general principles, validation,
  and sources of discrepancies.
\newblock {\em European Heart Journal-Cardiovascular Imaging}, 20(6):605--619,
  2019.

\bibitem{attar20193d}
Rahman Attar, Marco Perea{\~n}ez, Christopher Bowles, Stefan~K Piechnik, Stefan
  Neubauer, Steffen~E Petersen, and Alejandro~F Frangi.
\newblock 3d cardiac shape prediction with deep neural networks: Simultaneous
  use of images and patient metadata.
\newblock In {\em International Conference on Medical Image Computing and
  Computer-Assisted Intervention}, pages 586--594. Springer, 2019.

\bibitem{bai2015bi}
Wenjia Bai, Wenzhe Shi, Antonio de Marvao, Timothy~JW Dawes, Declan~P
  O’Regan, Stuart~A Cook, and Daniel Rueckert.
\newblock A bi-ventricular cardiac atlas built from 1000+ high resolution mr
  images of healthy subjects and an analysis of shape and motion.
\newblock {\em Medical image analysis}, 26(1):133--145, 2015.

\bibitem{Balakrishnan2019VoxelMorphAL}
Guha Balakrishnan, Amy Zhao, Mert~Rory Sabuncu, John~V. Guttag, and Adrian~V.
  Dalca.
\newblock Voxelmorph: A learning framework for deformable medical image
  registration.
\newblock {\em IEEE Transactions on Medical Imaging}, 38:1788--1800, 2019.

\bibitem{Beetz2021BiventricularSR}
Marcel Beetz, Abhirup Banerjee, and Vicente Grau.
\newblock Biventricular surface reconstruction from cine mri contours using
  point completion networks.
\newblock {\em 2021 IEEE 18th International Symposium on Biomedical Imaging
  (ISBI)}, pages 105--109, 2021.

\bibitem{bernard2018deep}
Olivier Bernard, Alain Lalande, Clement Zotti, Frederick Cervenansky, Xin Yang,
  Pheng-Ann Heng, Irem Cetin, Karim Lekadir, Oscar Camara, Miguel
  Angel~Gonzalez Ballester, et~al.
\newblock Deep learning techniques for automatic mri cardiac multi-structures
  segmentation and diagnosis: is the problem solved?
\newblock {\em IEEE transactions on medical imaging}, 37(11):2514--2525, 2018.

\bibitem{chen2020anatomy}
Pingjun Chen, Xiao Chen, Eric~Z Chen, Hanchao Yu, Terrence Chen, and Shanhui
  Sun.
\newblock Anatomy-aware cardiac motion estimation.
\newblock In {\em International Workshop on Machine Learning in Medical
  Imaging}, pages 150--159. Springer, 2020.

\bibitem{Chen2021ShapeRW}
Xiang Chen, Nishant Ravikumar, Yan Xia, Rahman Attar, Andr{\'e}s Diaz-Pinto,
  Stefan~K. Piechnik, Stefan Neubauer, Steffen~Erhard Petersen, and
  Alejandro~F. Frangi.
\newblock Shape registration with learned deformations for 3d shape
  reconstruction from sparse and incomplete point clouds.
\newblock {\em Medical image analysis}, 74:102228, 2021.

\bibitem{Corona2021SMPLicitTG}
Enric Corona, Albert Pumarola, G. Aleny{\`a}, Gerard Pons-Moll, and Francesc
  Moreno-Noguer.
\newblock Smplicit: Topology-aware generative model for clothed people.
\newblock {\em 2021 IEEE/CVF Conference on Computer Vision and Pattern
  Recognition (CVPR)}, pages 11870--11880, 2021.

\bibitem{Deng2021DeformedIF}
Yu Deng, Jiaolong Yang, and Xin Tong.
\newblock Deformed implicit field: Modeling 3d shapes with learned dense
  correspondence.
\newblock {\em 2021 IEEE/CVF Conference on Computer Vision and Pattern
  Recognition (CVPR)}, pages 10281--10291, 2021.

\bibitem{frangi2002automatic}
Alejandro~F Frangi, Daniel Rueckert, Julia~A Schnabel, and Wiro~J Niessen.
\newblock Automatic construction of multiple-object three-dimensional
  statistical shape models: Application to cardiac modeling.
\newblock {\em IEEE transactions on medical imaging}, 21(9):1151--1166, 2002.

\bibitem{Guo2021UnsupervisedLD}
Yuyu Guo, Lei Bi, Dongming Wei, Liyun Chen, Zhengbin Zhu, Dagan Feng, Ruiyan
  Zhang, Qian Wang, and Jinman Kim.
\newblock Unsupervised landmark detection based spatiotemporal motion
  estimation for 4d dynamic medical images.
\newblock {\em IEEE transactions on cybernetics}, PP, 2021.

\bibitem{Hao2020DualSDFSS}
Zekun Hao, Hadar Averbuch-Elor, Noah Snavely, and Serge~J. Belongie.
\newblock Dualsdf: Semantic shape manipulation using a two-level
  representation.
\newblock {\em 2020 IEEE/CVF Conference on Computer Vision and Pattern
  Recognition (CVPR)}, pages 7628--7638, 2020.

\bibitem{hoogendoorn2012high}
Corn{\'e} Hoogendoorn, Nicolas Duchateau, Damian Sanchez-Quintana, Tristan
  Whitmarsh, Federico~M Sukno, Mathieu De~Craene, Karim Lekadir, and
  Alejandro~F Frangi.
\newblock A high-resolution atlas and statistical model of the human heart from
  multislice ct.
\newblock {\em IEEE transactions on medical imaging}, 32(1):28--44, 2012.

\bibitem{Isensee2019nnUNetBT}
Fabian Isensee, Jens Petersen, Simon A.~A. Kohl, Paul~F. J{\"a}ger, and Klaus
  Maier-Hein.
\newblock nnu-net: Breaking the spell on successful medical image segmentation.
\newblock {\em ArXiv}, abs/1904.08128, 2019.

\bibitem{Jiang2022LoRDL4}
Boyan Jiang, Xinlin Ren, Mingsong Dou, Xiangyang Xue, Yanwei Fu, and Yinda
  Zhang.
\newblock Lord: Local 4d implicit representation for high-fidelity dynamic
  human modeling.
\newblock In {\em ECCV}, 2022.

\bibitem{Joyce2022RapidIO}
Thomas Joyce, Stefano Buoso, Christian~T. Stoeck, and Sebastian Kozerke.
\newblock Rapid inference of personalised left-ventricular meshes by
  deformation-based differentiable mesh voxelization.
\newblock {\em Medical image analysis}, 79:102445, 2022.

\bibitem{kong2021whole}
Fanwei Kong and Shawn~C Shadden.
\newblock Whole heart mesh generation for image-based computational simulations
  by learning free-from deformations.
\newblock In {\em International Conference on Medical Image Computing and
  Computer-Assisted Intervention}, pages 550--559. Springer, 2021.

\bibitem{Kong2021ADA}
Fanwei Kong, Nathan~M. Wilson, and Shawn~C. Shadden.
\newblock A deep-learning approach for direct whole-heart mesh reconstruction.
\newblock {\em Medical image analysis}, 74:102222, 2021.

\bibitem{krebs2019probabilistic}
Julian Krebs, Tommaso Mansi, Nicholas Ayache, and Herv{\'e} Delingette.
\newblock Probabilistic motion modeling from medical image sequences:
  application to cardiac cine-mri.
\newblock In {\em International Workshop on Statistical Atlases and
  Computational Models of the Heart}, pages 176--185. Springer, 2019.

\bibitem{lorensen1987marching}
William~E Lorensen and Harvey~E Cline.
\newblock Marching cubes: A high resolution 3d surface construction algorithm.
\newblock {\em ACM siggraph computer graphics}, 21(4):163--169, 1987.

\bibitem{mandikal20183d}
Priyanka Mandikal, KL Navaneet, Mayank Agarwal, and R~Venkatesh Babu.
\newblock 3d-lmnet: Latent embedding matching for accurate and diverse 3d point
  cloud reconstruction from a single image.
\newblock {\em arXiv preprint arXiv:1807.07796}, 2018.

\bibitem{meng2022mesh}
Qingjie Meng, Wenjia Bai, Tianrui Liu, Declan~P O’Regan, and Daniel Rueckert.
\newblock Mesh-based 3d motion tracking in cardiac mri using deep learning.
\newblock In {\em International Conference on Medical Image Computing and
  Computer-Assisted Intervention}, pages 248--258. Springer, 2022.

\bibitem{Meng2022MulViMotionS3}
Qingjie Meng, Chen Qin, Wenjia Bai, Tianrui Liu, Antonio de Marvao, Declan~P.
  O’Regan, and Daniel Rueckert.
\newblock Mulvimotion: Shape-aware 3d myocardial motion tracking from
  multi-view cardiac mri.
\newblock {\em IEEE Transactions on Medical Imaging}, 41:1961--1974, 2022.

\bibitem{Mescheder2019OccupancyNL}
Lars~M. Mescheder, Michael Oechsle, Michael Niemeyer, Sebastian Nowozin, and
  Andreas Geiger.
\newblock Occupancy networks: Learning 3d reconstruction in function space.
\newblock {\em 2019 IEEE/CVF Conference on Computer Vision and Pattern
  Recognition (CVPR)}, pages 4455--4465, 2019.

\bibitem{Mihajlovi2021LEAPLA}
Marko Mihajlovi{\'c}, Yan Zhang, Michael~J. Black, and Siyu Tang.
\newblock Leap: Learning articulated occupancy of people.
\newblock {\em 2021 IEEE/CVF Conference on Computer Vision and Pattern
  Recognition (CVPR)}, pages 10456--10466, 2021.

\bibitem{Niemeyer2019OccupancyF4}
Michael Niemeyer, Lars~M. Mescheder, Michael Oechsle, and Andreas Geiger.
\newblock Occupancy flow: 4d reconstruction by learning particle dynamics.
\newblock {\em 2019 IEEE/CVF International Conference on Computer Vision
  (ICCV)}, pages 5378--5388, 2019.

\bibitem{Palafox2021NPMsNP}
Pablo~Rodr{\'i}guez Palafox, Aljavz Bovzivc, Justus Thies, Matthias
  Nie{\ss}ner, and Angela Dai.
\newblock Npms: Neural parametric models for 3d deformable shapes.
\newblock {\em 2021 IEEE/CVF International Conference on Computer Vision
  (ICCV)}, pages 12675--12685, 2021.

\bibitem{park2019deepsdf}
Jeong~Joon Park, Peter Florence, Julian Straub, Richard Newcombe, and Steven
  Lovegrove.
\newblock Deepsdf: Learning continuous signed distance functions for shape
  representation.
\newblock In {\em Proceedings of the IEEE/CVF conference on computer vision and
  pattern recognition}, pages 165--174, 2019.

\bibitem{Park2019DeepSDFLC}
Jeong~Joon Park, Peter~R. Florence, Julian Straub, Richard~A. Newcombe, and S.
  Lovegrove.
\newblock Deepsdf: Learning continuous signed distance functions for shape
  representation.
\newblock {\em 2019 IEEE/CVF Conference on Computer Vision and Pattern
  Recognition (CVPR)}, pages 165--174, 2019.

\bibitem{qin2018joint}
Chen Qin, Wenjia Bai, Jo Schlemper, Steffen~E Petersen, Stefan~K Piechnik,
  Stefan Neubauer, and Daniel Rueckert.
\newblock Joint learning of motion estimation and segmentation for cardiac mr
  image sequences.
\newblock In {\em International Conference on Medical Image Computing and
  Computer-Assisted Intervention}, pages 472--480. Springer, 2018.

\bibitem{Qin2020BiomechanicsinformedNN}
Chen Qin, Shuo Wang, Chen Chen, Huaqi Qiu, Wenjia Bai, and Daniel Rueckert.
\newblock Biomechanics-informed neural networks for myocardial motion tracking
  in mri.
\newblock In {\em MICCAI}, 2020.

\bibitem{Raju2022DeepIS}
Ashwin Raju, Shun Miao, Chi-Tung Cheng, Le Lu, Mei Han, Jing Xiao, Chien-Hung
  Liao, Junzhou Huang, and Adam~P. Harrison.
\newblock Deep implicit statistical shape models for 3d medical image
  delineation.
\newblock {\em ArXiv}, abs/2104.02847, 2022.

\bibitem{Sun2022TopologyPreservingSR}
Shanlin Sun, Kun Han, Deying Kong, Hao Tang, Xiangyi Yan, and Xiaohui Xie.
\newblock Topology-preserving shape reconstruction and registration via neural
  diffeomorphic flow.
\newblock {\em 2022 IEEE/CVF Conference on Computer Vision and Pattern
  Recognition (CVPR)}, pages 20813--20823, 2022.

\bibitem{Tiwari2021NeuralGIFNG}
Garvita Tiwari, Nikolaos Sarafianos, Tony Tung, and Gerard Pons-Moll.
\newblock Neural-gif: Neural generalized implicit functions for animating
  people in clothing.
\newblock {\em 2021 IEEE/CVF International Conference on Computer Vision
  (ICCV)}, pages 11688--11698, 2021.

\bibitem{8809843}
Yifan Wang, Zichun Zhong, and Jing Hua.
\newblock Deeporgannet: On-the-fly reconstruction and visualization of 3d / 4d
  lung models from single-view projections by deep deformation network.
\newblock {\em IEEE Transactions on Visualization and Computer Graphics},
  26(1):960--970, 2020.

\bibitem{wickramasinghe2020voxel2mesh}
Udaranga Wickramasinghe, Edoardo Remelli, Graham Knott, and Pascal Fua.
\newblock Voxel2mesh: 3d mesh model generation from volumetric data.
\newblock In {\em International Conference on Medical Image Computing and
  Computer-Assisted Intervention}, pages 299--308. Springer, 2020.

\bibitem{Yang2022ImplicitAtlasLD}
Jiancheng Yang, Udaranga Wickramasinghe, Bingbing Ni, and P. Fua.
\newblock Implicitatlas: Learning deformable shape templates in medical
  imaging.
\newblock {\em 2022 IEEE/CVF Conference on Computer Vision and Pattern
  Recognition (CVPR)}, pages 15840--15850, 2022.

\bibitem{ye2020pc}
Meng Ye, Qiaoying Huang, Dong Yang, Pengxiang Wu, Jingru Yi, Leon Axel, and
  Dimitris Metaxas.
\newblock Pc-u net: Learning to jointly reconstruct and segment the cardiac
  walls in 3d from ct data.
\newblock In {\em International Workshop on Statistical Atlases and
  Computational Models of the Heart}, pages 117--126. Springer, 2020.

\bibitem{Ye2021DeepTagAU}
Meng Ye, Mikael Kanski, D. Yang, Qi Chang, Zhennan Yan, Qiaoying Huang, Leon
  Axel, and Dimitris~N. Metaxas.
\newblock Deeptag: An unsupervised deep learning method for motion tracking on
  cardiac tagging magnetic resonance images.
\newblock {\em 2021 IEEE/CVF Conference on Computer Vision and Pattern
  Recognition (CVPR)}, pages 7257--7267, 2021.

\bibitem{Yu2020MotionPN}
Hanchao Yu, Xiao Chen, Humphrey Shi, Terrence Chen, Thomas~S. Huang, and
  Shanhui Sun.
\newblock Motion pyramid networks for accurate and efficient cardiac motion
  estimation.
\newblock {\em ArXiv}, abs/2006.15710, 2020.

\bibitem{yu2020motion}
Hanchao Yu, Xiao Chen, Humphrey Shi, Terrence Chen, Thomas~S Huang, and Shanhui
  Sun.
\newblock Motion pyramid networks for accurate and efficient cardiac motion
  estimation.
\newblock In {\em International Conference on Medical Image Computing and
  Computer-Assisted Intervention}, pages 436--446. Springer, 2020.

\bibitem{yu2020foal}
Hanchao Yu, Shanhui Sun, Haichao Yu, Xiao Chen, Honghui Shi, Thomas~S Huang,
  and Terrence Chen.
\newblock Foal: Fast online adaptive learning for cardiac motion estimation.
\newblock In {\em Proceedings of the IEEE/CVF Conference on Computer Vision and
  Pattern Recognition}, pages 4313--4323, 2020.

\bibitem{Zhang2022LearningCO}
Xiaoran Zhang, Chenyu You, Shawn~S. Ahn, Juntang Zhuang, Lawrence~H. Staib, and
  James~S. Duncan.
\newblock Learning correspondences of cardiac motion from images using
  biomechanics-informed modeling.
\newblock {\em ArXiv}, abs/2209.00726, 2022.

\bibitem{Zheng2021DeepIT}
Zerong Zheng, Tao Yu, Qionghai Dai, and Yebin Liu.
\newblock Deep implicit templates for 3d shape representation.
\newblock {\em 2021 IEEE/CVF Conference on Computer Vision and Pattern
  Recognition (CVPR)}, pages 1429--1439, 2021.

\bibitem{zhou2019one}
Xiao-Yun Zhou, Zhao-Yang Wang, Peichao Li, Jian-Qing Zheng, and Guang-Zhong
  Yang.
\newblock One-stage shape instantiation from a single 2d image to 3d point
  cloud.
\newblock In {\em International Conference on Medical Image Computing and
  Computer-Assisted Intervention}, pages 30--38. Springer, 2019.

\end{thebibliography}
